\documentclass{article}

     \PassOptionsToPackage{numbers, compress}{natbib}

\usepackage[final]{neurips_data_2021}

\usepackage[utf8]{inputenc} 
\usepackage[T1]{fontenc}    
\usepackage{hyperref}       
\usepackage{multirow}
\usepackage{url}            
\usepackage{booktabs}       
\usepackage{amsfonts}       
\usepackage{nicefrac}       
\usepackage{microtype}      
\usepackage{xcolor}         
\usepackage{graphicx}         %
\usepackage{subcaption}
\usepackage[arabic, english]{babel}  
\usepackage{adjustbox}
\usepackage{wrapfig}
\usepackage{textcomp}
\usepackage[compact]{titlesec}
\usepackage{diagbox}
\usepackage{xspace}

\newcommand{\corpusname}{LSH methods for data deduplication in a Wikipedia   artificial dataset\xspace}

\title{\corpusname}

\author{%
Juan Ciro\\
Factored\\
\And
Daniel Galvez\\
NVIDIA\\
\And
Tim Schlippe\\
IU University of Applied Sciences\\
\And
David Kanter\\
MLCommons\\
}

\begin{document}

\maketitle

\begin{abstract}
This paper illustrates locality sensitive hasing (LSH) models for the identification and removal of nearly redundant data in a text dataset. To evaluate the different models, we create an artificial dataset for data deduplication using English Wikipedia articles. Area-Under-Curve (AUC) over 0.9 were observed for most models, with the best model reaching 0.96. Deduplication enables more effective model training by preventing the model from learning a distribution that differs from the real one as a result of the repeated data.
\end{abstract}

\section{Introduction}

The introduction of large data sets for deep learning models has had a significant impact on NLP~\cite{brown2020language, dodge2021documenting} and ASR~\cite{galvez2021people, mazumder2021multilingual}. This expansion has created substantial issues for data quality assurance, as performing manual examination on millions of data points becomes unfeasible.

Duplicate and nearly duplicate data are one of the most significant issues with these collections. The usage of several data sources can cause the same file to be incorporated into training data sets multiple times, since data is easily copied on the Internet. The main challenge is not the same exact data, because there are easy and efficient approaches for detecting and eliminating it (e.g., hashing every document in the data set to an integer and checking for collisions). The key issue is in detecting data which is practically the same, though not exactly the same, i.e. examples with few changes, deletions, or additions of sentences. Data duplication is a critical issue for two reasons:

\begin{itemize}
\item  Failure to guarantee that no duplicate data exists may result in overfitting in a specific section of the distribution or an unfair bias in the model ~\cite{gao_making_2021}.

\item 
Removing duplicated data also allows us to partition our data set into train, dev, and test parts while guaranteeing that they have no data overlap.

\end{itemize}

\section{Related Work}

Despite the importance of data deduplication, there is no standard way to do it. For example, in~\cite{radford2019language},  a followup study was conducted on several tasks in which the GPT-2 model was validated; using bloom filters, 8-gram overlaps between the training and validation sets were identified. In the case of GPT-3, it was decided to discard this data prior to evaluating it using the 13-gram overlap approach~\cite{brown2020language, trinh2019simple}. A different approach was applied in~\cite{dodge2021documenting} in which they performed perfect match after verbatim processing to discover an average of 3 percent matches between the C4 dataset and a variety of other validation datasets. A model based on perplexity is presented in~\cite{gao_making_2021}.

\section{Constructing the dataset and the model}
\label{sec:construction}

\subsection{Dataset Construction}

The evaluation of data deduplication methods is difficult due to the lack of a dedicated dataset. Using over 30,000 distinct Wikipedia articles of varying lengths, we create artificial, nearly duplicated samples. To achieve this we tokenize the texts' sentences. After that, we randomly select sentences for removal. Then some sentences were added by taking 2,000 different sentences collected from 300 articles that were not included in the final dataset with a length of 8 to 20 words. Between 87 and 94 percent resemblance was preserved between the original text and the synthetic duplicate.
With this method, we were able to create a 6,258 artificial articles created from the original 24,999. Obtaining in total a database with 31,257 texts, including duplicates and originals, 6,258 labeled as duplicates and 24,999 as non-duplicates.We also construct a test data set of 6,286 articles using the same methodology to validate our results, 1,258 labeled as duplicates and 5,028 as non-duplicates.

\subsection{Model}
One purpose of this project is to create methods for large datasets, hence the model must be scalable. The model used is based on Local Sensitive Hashing (LSH) ~\cite{zhu2016LSH}. The model is known as the classic LSH technique or MinHash, which groups the samples into buckets after being encoded using Minhash functions, where each sample in a bucket is likely to have a low Jaccard Distance, as measured by character n-grams~\cite{datasketch:online, jafari2021survey}. Finally, members of the same bucket are compared using Jaccard similarity; if the Jaccard similarity exceeds the threshold hyperparameter, two documents are considered "the same". The ability to compare samples only to samples that fall into the same bucket makes LSH $O(N)$ to evaluate, as opposed to $O(N^2)$ (which applies to most k-nearest-neighbors approaches) on a dataset of N points~\cite{jafari2020experimental, lei2020locality}.

\subsection{Metrics}
The models were evaluated using two key measures: speed and AUC. AUC was used instead of accuracy, since finding duplicate data is necessary but not at the expense of eliminating good data.

\label{subsec:extraction}

\section{Tuning and Evaluating} \label{sec: eval}

\subsection{Model Tuning}

The model has three hyperparameters: the number of hashing functions, the size of each character n-gram, and the threshold for classifying an example as duplicate. Despite the fact that there are superior combinations of values in the training results, none of the combinations produced an AUC below 0.8. This means that while the choice of hyperparameters is pertinent and may vary depending on the dataset, the method has broad applicability.

\subsection{Evaluation}

The model was evaluated using the previously specified testing dataset and the three models with the greatest AUC. The evaluation findings showed the generalization of the models, since none of the 3
models utilized show a significant drop in AUC values. Results in Table~\ref{tab:results}.

\begin{table}[t!]
\centering
\caption{Results for the different models}
{
 \begin{tabular}{c c c c c c}
\toprule
Number permutation & Threshold & N-gram &  AUC  & Speed inference test (minutes) \\ \midrule

128 & 0.65 & 12 & 0.96 & 12.21 \\
128 & 0.80 & 16 & 0.94 & 12.64 \\
64 & 0.75 & 12 & 0.91 & 11.13 \\

\bottomrule
\end{tabular}
}

\label{tab:results}
\end{table}

\section{Conclusion}

Tools that ensure the data's quality will become highly significant. Data duplication is a common issue that arises during the development of big datasets. We provide a model in this work that enables the elimination of this data. Additionally, the establishment of a dataset for measuring their performance was discussed. The complete source code is available~\cite{peoplesspeech_code}.

\begin{table}[t!]
\centering
\caption{Results for the different models}
{
 \begin{tabular}{c c c }
\toprule
 Dataset & \% Duplicated  \\ \midrule

PTB & 2.67\% \\
WikiText-2 & 0.66\% \\
Enwik8 & 7.50\% \\
Text8 & 2.34\% \\
WikiText-3 & 9.09\% \\
1BW & 13.19\% \\

\bottomrule
\end{tabular}
}

\label{tab:results}
\end{table}

\bibliography{msc}
\bibliographystyle{abbrvnat}

\end{document}